\documentclass[]{fairmeta}

\usepackage{wrapfig}
\usepackage{tabularx}

\def\vx{{\bm{x}}}

\newlength\savewidth
\newcommand{\tablestyle}[2]{\setlength{\tabcolsep}{#1}\renewcommand{\arraystretch}{#2}\centering\footnotesize}

\newcolumntype{x}[1]{>{\centering\arraybackslash}p{#1pt}}
\newcolumntype{y}[1]{>{\raggedright\arraybackslash}p{#1pt}}
\newcolumntype{z}[1]{>{\raggedleft\arraybackslash}p{#1pt}}

\renewcommand{\paragraph}[1]{\vspace{1.25mm}\noindent\textbf{#1}}

\usepackage{algorithm}
\usepackage{listings}

\definecolor{codeblue}{rgb}{0.25, 0.5, 0.5}
\definecolor{codekw}{rgb}{0.35, 0.35, 0.75}
\lstdefinestyle{Pytorch}{
    language = Python,
    backgroundcolor = \color{white},
    basicstyle = \fontsize{9pt}{8pt}\selectfont\ttfamily\bfseries,
    columns = fullflexible,
    aboveskip=1pt,
    belowskip=1pt,
    breaklines = true,
    captionpos = b,
    commentstyle = \color{codeblue},
    keywordstyle = \color{codekw},
}

\definecolor{green}{HTML}{009000}
\definecolor{red}{HTML}{ea4335}
\newcommand{\better}[1]{\textcolor{green}{$\uparrow\,$#1}}
\newcommand{\worse}[1]{\textcolor{red}{$\downarrow\,$#1}}
\newcommand{\betterinv}[1]{\textcolor{green}{$\downarrow\,$#1}}
\newcommand{\worseinv}[1]{\textcolor{red}{$\uparrow\,$#1}}

\title{Transformers without Normalization}

\author[1,2]{Jiachen Zhu}
\author[1]{Xinlei Chen}
\author[3]{Kaiming He}
\author[1,2]{Yann LeCun}
\author[1,4,\dagger]{Zhuang Liu}

\affiliation[1]{FAIR, Meta}
\affiliation[2]{New York University}
\affiliation[3]{MIT}
\affiliation[4]{Princeton University}
\contribution[\dagger]{Project lead}

\abstract{
Normalization layers are ubiquitous in modern neural networks and have long been considered essential.
This work demonstrates that Transformers without normalization can achieve the same or better performance using a remarkably simple technique.
We introduce Dynamic Tanh (DyT), an element-wise operation $\mathrm{DyT}(\vx) = \tanh(\alpha \vx)$, as a drop-in replacement for normalization layers in Transformers.
DyT is inspired by the observation that layer normalization in Transformers often produces tanh-like, $S$-shaped input-output mappings.
By incorporating DyT, Transformers without normalization can match or exceed the performance of their normalized counterparts, mostly without hyperparameter tuning.
We validate the effectiveness of Transformers with DyT across diverse settings, ranging from recognition to generation, supervised to self-supervised learning, and computer vision to language models.
These findings challenge the conventional understanding that normalization layers are indispensable in modern neural networks, and offer new insights into their role in deep networks.
}

\date{\today} 
\metadata[Project page and code]{{\href{https://jiachenzhu.github.io/DyT}{\texttt{jiachenzhu.github.io/DyT}}}}
\metadata[Correspondence]{\email{jiachen.zhu@nyu.edu}, \email{zhuangl@princeton.edu}}
\begin{document}

\maketitle

\section{Introduction}

Over the past decade, normalization layers have solidified their positions as one of the most fundamental components of modern neural networks.
It all traces back to the invention of batch normalization in 2015~\citep{ioffe2015batch}, which enabled drastically faster and better convergence in visual recognition models and quickly gained momentum in the following years.
Since then, many variants of normalization layers have been proposed for different network architectures or domains~\citep{ba2016layer,ulyanov2016instance,wu2018group,zhang2019root}.
Today, virtually all modern networks use normalization layers, with layer normalization (Layer Norm, or LN)~\citep{ba2016layer} being one of the most popular, particularly in the dominant Transformer architecture~\citep{vaswani2017attention, dosovitskiy2020image}.

The widespread adoption of normalization layers is largely driven by their empirical benefits in optimization \citep{santurkar2018does, bjorck2018understanding}.
In addition to achieving better results, they help accelerate and stabilize convergence.
As neural networks become wider and deeper, this necessity becomes ever more critical \citep{brock2021characterizing, huang2023normalization}.
Consequently, normalization layers are widely regarded as crucial, if not indispensable, for the effective training of deep networks.
This belief is subtly evidenced by the fact that, in recent years, novel architectures often seek to replace attention or convolution layers \citep{tolstikhin2021mlp,gu2023mamba, sun2024learning, feng2024were}, but almost always retain the normalization layers.

This paper challenges this belief by introducing a simple alternative to normalization layers in Transformers.
Our exploration starts with the observation that LN layers map their inputs to outputs with tanh-like, $S$-shaped curves,  scaling the input activations while squashing the extreme values.
Inspired by this insight, we propose an element-wise operation termed Dynamic Tanh (DyT), defined as: $\mathrm{DyT}(\vx) = \tanh(\alpha \vx)$, where $\alpha$ is a learnable parameter.
This operation aims to emulate the behavior of LN by learning an appropriate scaling factor through $\alpha$ and squashing extreme values via the bounded tanh function.
Notably, unlike normalization layers, it achieves both effects without the need to compute activation statistics.

\begin{figure}[t]
\centering
\vspace{0.4in}
\includegraphics[width=0.7\linewidth]{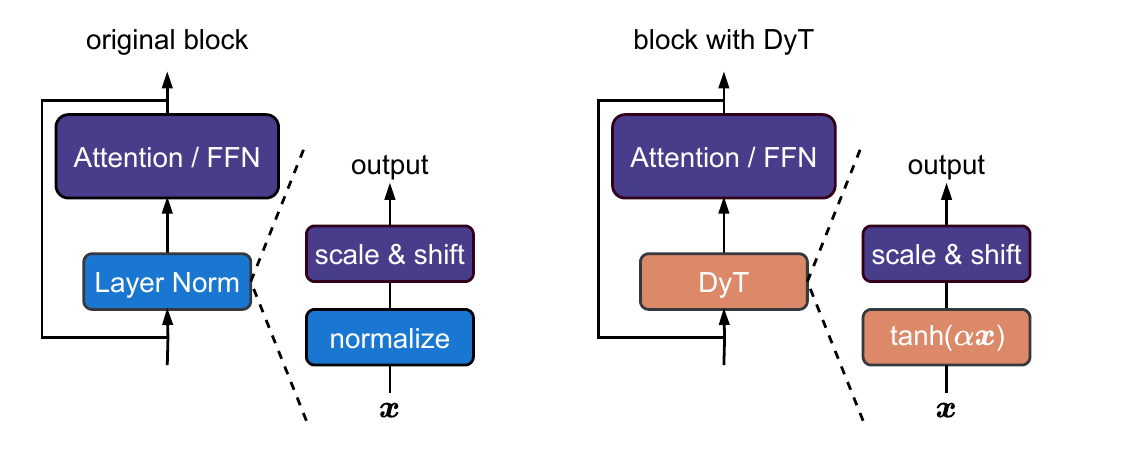}
\caption{\emph{Left:} original Transformer block. \emph{Right:} block with our proposed Dynamic Tanh (DyT) layer. DyT is a straightforward replacement for commonly used Layer Norm~\citep{ba2016layer} (in some cases RMSNorm~\citep{zhang2019root}) layers. Transformers with DyT match or exceed the performance of their normalized counterparts.}
\label{figure:before_after}
\end{figure}

Employing DyT is straightforward, as shown in Figure~\ref{figure:before_after}: we directly replace existing normalization layers with DyT in architectures such as vision and language Transformers. We empirically demonstrate that models with DyT can train stably and achieve high final performance across a wide range of settings.
It often does not require tuning the training hyperparameters on the original architecture. 
Our work challenges the notion that normalization layers are indispensable for training modern neural networks and provides empirical insights into the properties of normalization layers.

\section{Background: Normalization Layers}

We begin by reviewing the normalization layers. 
Most normalization layers share a common formulation.
Given an input $\vx$ with shape $(B, T, C)$, where $B$ is the batch size, $T$ is the number of tokens, and $C$ is the embedding dimension per token, the output is generally computed as: 
\begin{equation}
\mathrm{normalization}(\vx) = \bm \gamma * \left ( \frac{\vx - \bm \mu}{\sqrt{\bm \sigma^2 + \epsilon }}  \right ) + \bm \beta
\end{equation}
where $\epsilon$ is a small constant, and $\bm \gamma$ and $\bm \beta$ are learnable vector parameters of shape $(C,)$. They are ``scaling'' and ``shifting'' affine parameters that allow the output to be in any range.
The terms $\bm \mu$ and $\bm \sigma^2$ denote the mean and variance of the input. Different methods mainly differ in how these two statistics are computed. This results in $\bm \mu$ and $\bm \sigma^2$ having different dimensions, each with broadcasting applied during computation.

Batch normalization (BN)~\citep{ioffe2015batch} is the first modern normalization layer, and it has been primarily used in ConvNet models~\citep{szegedy2016rethinking,he2016deep,xie2017aggregated}. Its introduction represents a major milestone in deep learning architecture designs. BN computes the mean and variance across both the batch and token dimensions, specifically: $\mu_k = \frac{1}{B T} \sum_{i,j} x_{i j k}$ and $\sigma^2_k = \frac{1}{B T} \sum_{i, j} \left( x_{i j k} - \mu_k \right)^2 $. Other normalization layers popular in ConvNets, such as group normalization \citep{wu2018group} and instance normalization \citep{ulyanov2016instance}, were initially proposed for specialized tasks such as object detection and image stylization. They share the same overall formulation but differ in the axes and ranges over which the statistics are computed.

Layer normalization (LN)~\citep{ba2016layer} and 
root mean square normalization (RMSNorm)~\citep{zhang2019root} are the major two types of normalization layers used in Transformer architectures.
LN computes these statistics independently for each token in each sample, where $\mu_{i j} = \frac{1}{C} \sum_{k} x_{i j k}$ and $\sigma^2_{i j} = \frac{1}{C} \sum_{k} \left( x_{i j k} - \mu_{ij} \right)^2 $.
RMSNorm \citep{zhang2019root} simplifies LN by removing the mean-centering step and normalizing the input with $\mu_{ij} = 0$ and $\sigma^2_{i j} = \frac{1}{C} \sum_{k} x^2_{i j k}$. 
Today, most modern neural networks use LN due to its simplicity and universality. Recently, RMSNorm has gained popularity, particularly in language models like T5~\citep{raffel2020exploring}, LLaMA~\citep{touvron2023llama,touvron2023llama2, dubey2024llama}, Mistral~\citep{jiang2023mistral}, Qwen~\citep{bai2023qwen, yang2024qwen2}, InternLM~\citep{zhang2024internlm, cai2024internlm2} and DeepSeek \citep{liu2024deepseek, guo2025deepseek}. The Transformers we examine in this work all use LN, except that LLaMA uses RMSNorm.

\section{What Do Normalization Layers Do?}

\paragraph{Analysis setup.}
We first empirically study the behaviors of normalization layers in trained networks. For this analysis, we take a Vision Transformer model (ViT-B) \citep{dosovitskiy2020image} trained on ImageNet-1K \citep{deng2009imagenet}, a wav2vec 2.0 Large Transformer model~\citep{baevski2020wav2vec} trained on LibriSpeech~\citep{panayotov2015librispeech}, and a Diffusion Transformer (DiT-XL)~\citep{peebles2023scalable} trained on ImageNet-1K. In all cases, LN is applied in every Transformer block and before the final linear projection.

\begin{figure*}[t]
\centering
\begin{minipage}{\textwidth}
\hspace*{-0.5cm}
\centering
\includegraphics[width=\textwidth]{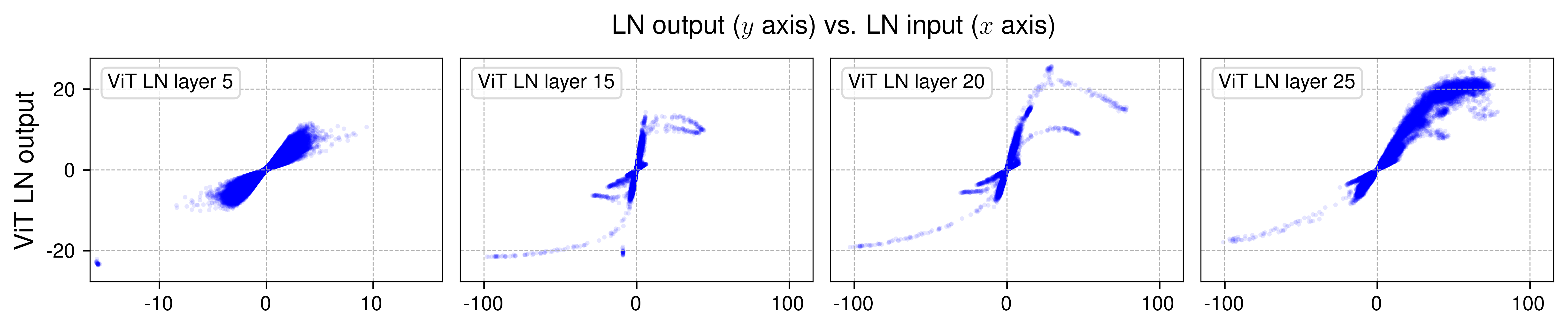}
\end{minipage}
\begin{minipage}{\textwidth}
\vspace*{-0.2cm}
\hspace*{-0.5cm}
  \centering
  \includegraphics[width=\textwidth]{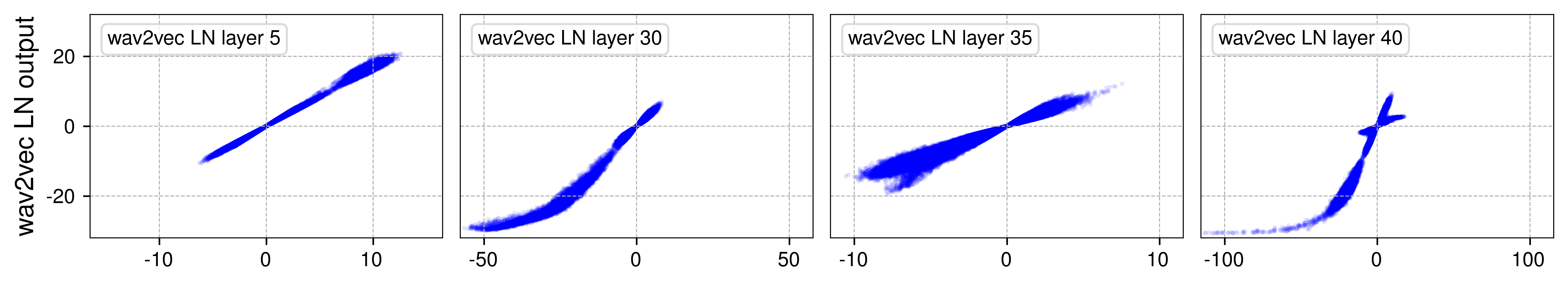}
\end{minipage}
\begin{minipage}{\textwidth}
\vspace*{-0.2cm}
\hspace*{-0.5cm}
  \centering
  \includegraphics[width=\textwidth]{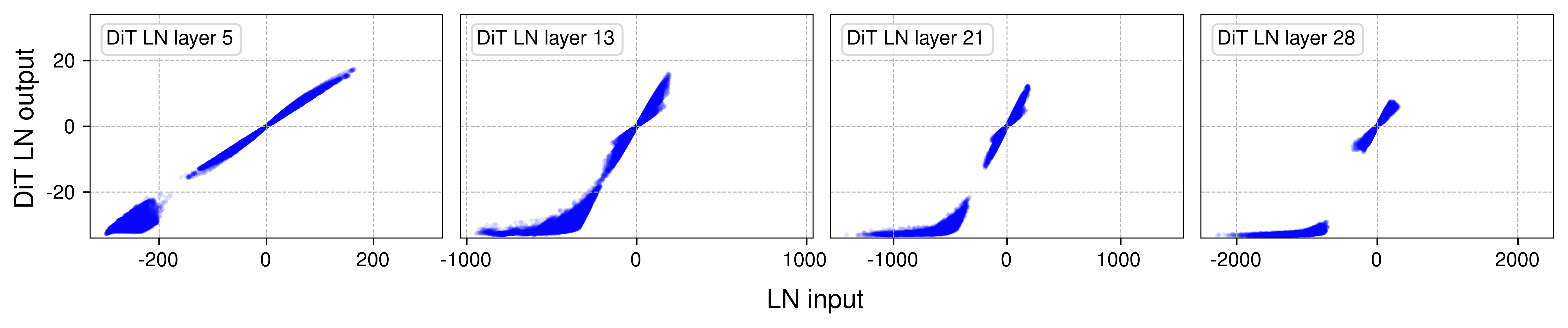}
\end{minipage}
\caption{\textbf{Output vs. input of selected layer normalization (LN) layers in Vision Transformer (ViT)~\citep{dosovitskiy2020image}, wav2vec 2.0 (a Transformer model for speech)~\citep{baevski2020wav2vec}, and Diffusion Transformer (DiT)~\citep{peebles2023scalable}.} 
We sample a mini-batch of samples and plot the input / output values of four LN layers in each model. The outputs are before the affine transformation in LN. The $S$-shaped curves highly resemble that of a tanh function (see Figure~\ref{figure:tanh_plot}). The more linear shapes in earlier layers can also be captured by the center part of a tanh curve. This motivates us to propose Dynamic Tanh (DyT) as a replacement, with a learnable scaler $\alpha$ to account for different scales on the $x$ axis.}
\label{figure:input-output}
\end{figure*}

For all three trained networks, we sample a mini-batch of samples and do a forward pass through the network. We then measure the input and output for the normalization layers, i.e., tensors immediately before and after the normalization operation, before the learnable affine transformation.
Since LN preserves the dimensions of the input tensor, we can establish a one-to-one correspondence between the input and output tensor elements, allowing for a direct visualization of their relationship. We plot the resulting mappings in Figure~\ref{figure:input-output}.

\begin{wrapfigure}[10]{r}{0.39\textwidth}
\centering
\vspace*{-0.6cm}
\hspace*{-0.8cm}
\includegraphics[width=1.15\linewidth]{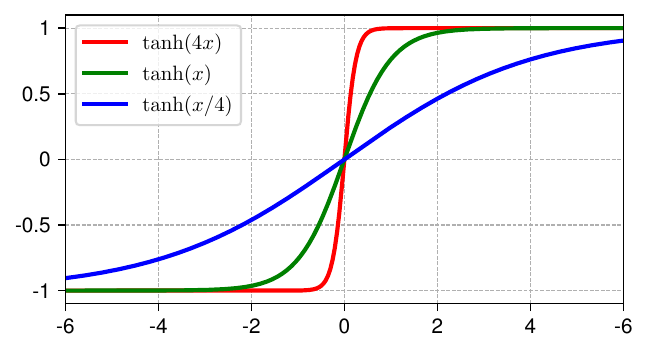}
\caption{$\tanh(\alpha x)$ with three different $\alpha$ values.}
\label{figure:tanh_plot}
\end{wrapfigure}

\paragraph{Tanh-like mappings with layer normalization.} For all three models, in earlier LN layers (1st column of Figure~\ref{figure:input-output}), we find this input-output relationship to be mostly linear, resembling a straight line in an $x$-$y$ plot. However, the deeper LN layers are places where we make more intriguing observations.

A striking observation from these deeper layers is that most of these curves' shapes highly resemble full or partial $S$-shaped curves represented by a tanh function (see Figure~\ref{figure:tanh_plot}). One might expect LN layers to linearly transform the input tensor, as subtracting the mean and dividing by standard deviation are linear operations. LN normalizes in a per-token manner, only linearly transforming each token's activations. As tokens have different mean and standard deviation values, the linearity does not hold collectively on all activations of the input tensor.
Nonetheless, it is still surprising to us that the actual non-linear transformation is highly similar to a scaled tanh function.

\vskip 0.2cm
For such an $S$-shaped curve, we note that the central part, represented by points with $x$ values close to zero, is still mainly in a linear shape. Most points ($\sim$99\%) fall in this linear range. 
However, there are still many points that clearly fall out of this range, which are considered to have ``extreme'' values, e.g., those with $x$ larger than 50 or smaller than -50 in the ViT model. Normalization layers' main effect for these values is to \emph{squash} them into less extreme values, more in line with the majority of points. This is where normalization layers could not approximated by a simple affine transformation layer. We hypothesize this non-linear and disproportional squashing effect on extreme values is what makes normalization layers important and indispensable. 

\vskip 0.2cm
Recent findings by \citet{ni2024nonlinearity} similarly highlight the strong non-linearities introduced by LN layers, demonstrating how the non-linearity enhances a model’s representational capacity. Moreover, this squashing behavior mirrors the saturation properties of biological neurons for large inputs, a phenomenon first observed about a century ago \citep{adrian1926impulses, adrian1926impulses2, adrian1926impulses3}.

\begin{figure*}[t]
\centering
\vspace{12ex}
\includegraphics[width=0.98\textwidth]{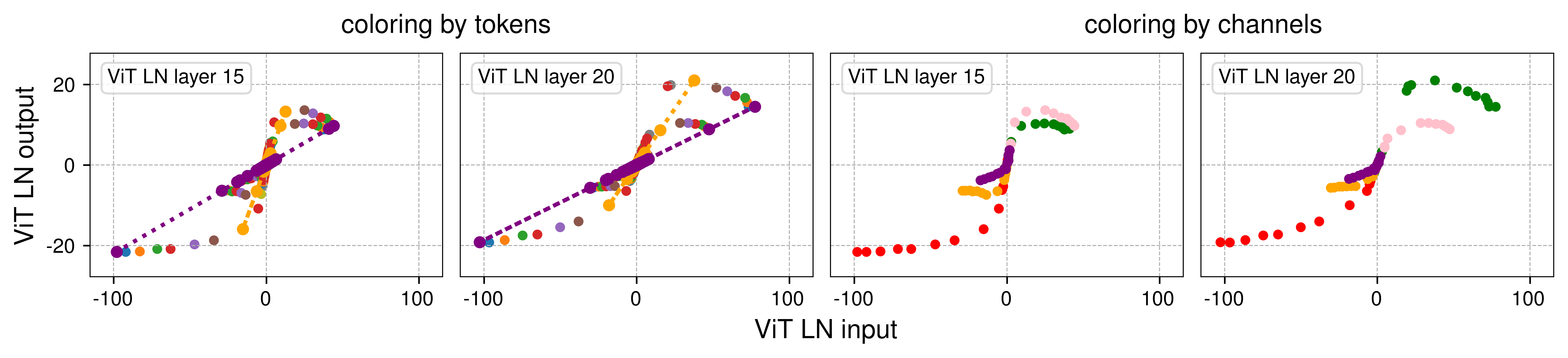}
\vspace{2pt}
\caption{\textbf{Output vs. input of two LN layers, with tensor elements colored to indicate different channel and token dimensions.} The input tensor has a shape of (samples, tokens, and channels), with elements visualized by assigning consistent colors to the same tokens (left two panels) and channels (right two panels). \emph{Left two panels}: points representing the same token (same color) form straight lines across different channels, as LN operates linearly across channels for each token. Interestingly, when plotted collectively, these lines form a non-linear tanh-shaped curve. 
\emph{Right two panels}: each channel's input spans different ranges on the $x$-axis, contributing distinct segments to the overall tanh-shaped curve. Certain channels (e.g., red, green, and pink) exhibit more extreme $x$ values, which are squashed by LN.}
\label{figure:input-output-color}
\end{figure*}

\vskip 0.2cm
\paragraph{Normalization by tokens and channels.} How does an LN layer perform a linear transformation for each token but also squash the extreme values in such a non-linear fashion?
To understand this, we visualize the points grouped by tokens and channels, respectively. This is plotted in Figure~\ref{figure:input-output-color} by taking the second and third subplots for ViT from Figure~\ref{figure:input-output}, but with a sampled subset of points for more clarity. When we select the channels to plot, we make sure to include the channels with extreme values. 

\vskip 0.2cm
On the left two panels of Figure~\ref{figure:input-output-color}, we visualize each token's activations using the same color. We observe that all points from any single token do form a straight line. However, since each token has a different variance, the slopes are different. Tokens with smaller input $x$ ranges tend to have smaller variance, and the normalization layer will divide their activations using a smaller standard deviation, hence producing a larger slope in the straight line. Collectively, they form an $S$-shaped curve that resembles a tanh function.
In the two panels on the right, we color each channel's activations using the same color. We find that different channels tend to have drastically different input ranges, with only a few channels (e.g., red, green, and pink) exhibiting large extreme values. These are the channels that get squashed the most by the normalization layer.


\section{Dynamic Tanh (DyT)}

Inspired by the similarity between the shapes of normalization layers and a scaled tanh function, we propose Dynamic Tanh (DyT) as a drop-in replacement for normalization layers. Given an input tensor $\vx$, a DyT layer is defined as follows:
\begin{equation}
\mathrm{DyT}(\vx) = \bm{\gamma} * \tanh(\alpha \vx) + \bm{\beta}
\end{equation}
where $\alpha$ is a learnable scalar parameter that allows scaling the input differently based on its range, accounting for varying $x$ scales (Figure~\ref{figure:input-output}). This is also why we name the whole operation ``Dynamic'' Tanh.
$\bm{\gamma}$ and $\bm{\beta}$ are learnable, per-channel vector parameters, the same as those used in all normalization layers---they allow the output to scale back to any scales. This is sometimes considered a separate affine layer; for our purposes, we consider them to be part of the DyT layer, just like how normalization layers also include them. See Algorithm~\ref{algorithm:dyt-pytorch} for implementation of DyT in Pytorch-like pseudocode.

\begin{wrapfigure}[12]{t}{0.45\textwidth}
\centering
\begin{minipage}{\linewidth}
\vspace*{-0.65cm}
\begin{algorithm}[H]
\begin{lstlisting}[style=Pytorch,escapeinside={(@}{@)}]
# input x has the shape of [B, T, C]
# B: batch size, T: tokens, C: dimension 

class DyT(Module):
    def __init__(self, C, init_(@$\bm \alpha$@)):
        super().__init__()
        self.(@$\bm \alpha$@) = Parameter(ones(1) * init_(@$\bm \alpha$@))
        self.(@$\bm \gamma$@) = Parameter(ones(C))
        self.(@$\bm \beta$@) = Parameter(zeros(C))

    def forward(self, x):
        x = tanh(self.alpha * x)
        return self.(@$\bm \gamma$@) * x + self.(@$\bm \beta$@)
        
\end{lstlisting}
\caption{Pseudocode of DyT layer.}
\label{algorithm:dyt-pytorch}
\end{algorithm}
\end{minipage}
\end{wrapfigure}
Integrating DyT layers into an existing architecture is straightforward: one DyT layer replaces one normalization layer (see Figure~\ref{figure:before_after}). This applies to normalization layers within attention blocks, FFN blocks, and the final normalization layer. Although DyT may look like or be considered an activation function, this study only uses it to replace normalization layers without altering any parts of the activation functions in the original architectures, such as GELU or ReLU.
Readers interested in the use of harmonic or hyperbolic functions as activation functions can refer to~\citet{hashemi2024can}. We also observe that there is little need to tune the hyperparameters used by the original architectures for DyT to perform well.

\vskip 0.2cm
\paragraph{On scaling parameters.} We always simply initialize $\bm{\gamma}$ to an all-one vector and $\bm{\beta}$ to an all-zero vector following normalization layers. For the scaler parameter $\alpha$, a default initialization of 0.5 is generally sufficient, except for LLM training.
A detailed analysis of $\alpha$ initialization is provided in Section \ref{section:alpha_init}.
Unless explicitly stated otherwise, $\alpha$ is initialized to 0.5 in our subsequent experiments.

\vskip 0.2cm
\paragraph{Remarks.} DyT is \emph{not} a new type of normalization layer, as it operates on each input element from a tensor independently during a forward pass without computing statistics or other types of aggregation. It does, however, preserve the effect of normalization layers in squashing the extreme values in a non-linear fashion while almost linearly transforming the very central parts of the input.

\section{Experiments}
\label{section:experiments}

To demonstrate the effectiveness of DyT, we experiment with Transformers and a few other modern architectures across a diverse range of tasks and domains.
In each experiment, we replace the LN or RMSNorm in the original architectures with DyT layers and follow the official open-source protocols to train and test both versions of the models. Detailed instructions for reproducing our results are provided in Appendix \ref{section:reproduce}.
Notably, to highlight the simplicity of adapting DyT, we use hyperparameters identical to those utilized by the normalized counterparts.
For completeness, additional experimental results regarding tuning of learning rates and initial values of $\alpha$ are provided in Appendix \ref{section:tuning}.

\begin{figure*}[t]
\vspace{5ex}
\centering
\begin{minipage}{0.49\textwidth}
\includegraphics[width=\textwidth]{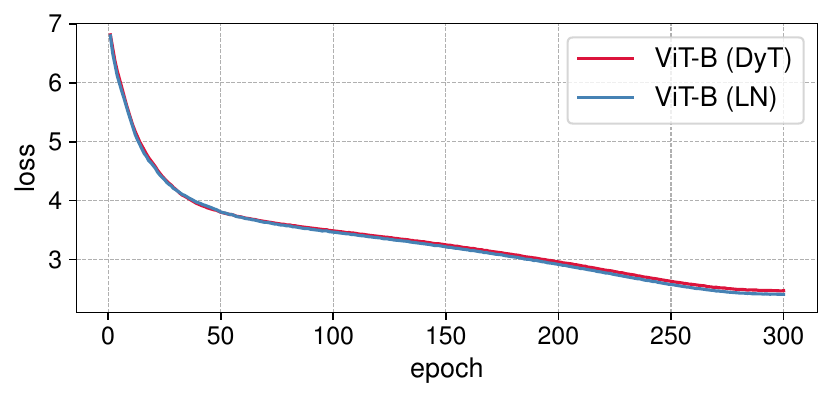}
\end{minipage}
\hfill
\begin{minipage}{0.49\textwidth}
\hspace*{-0.3cm}
  \includegraphics[width=\textwidth]{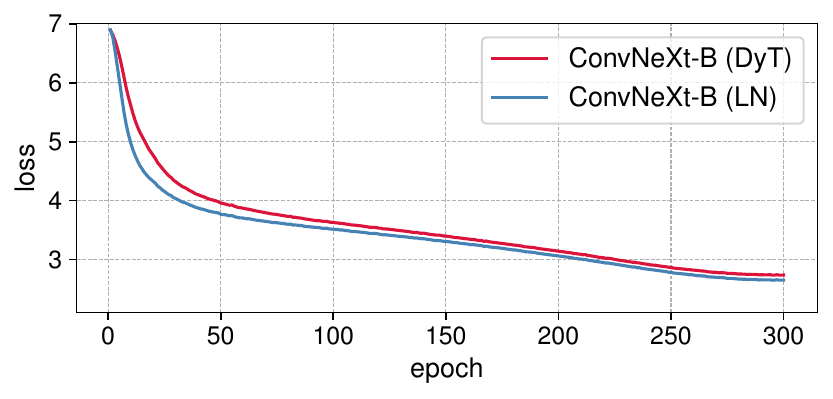}
\end{minipage}
\caption{\textbf{Training loss curves for ViT-B and ConvNeXt-B models.} The loss curves for both model types exhibit similar patterns between LN and DyT, suggesting that LN and DyT may share similar learning dynamics.}
\label{figure:sup_curve}
\vspace{2ex}
\end{figure*}

\paragraph{Supervised learning in vision.}
We train Vision Transformer (ViT) \citep{dosovitskiy2020image} and ConvNeXt \citep{liu2022convnet} of ``Base'' and ``Large'' sizes on the ImageNet-1K classification task \citep{deng2009imagenet}.
These models are selected due to their popularity and distinct operations: attention in ViT and convolution in ConvNeXt.
Table~\ref{table:classification} reports the top-1 classification accuracies.
DyT performs slightly better than LN across both architectures and model sizes.
We further plot the training loss for ViT-B and ConvNeXt-B in Figure~\ref{figure:sup_curve}. The curves show that the convergence behaviors of DyT and LN-based models are highly aligned.

\begin{table}[h]
\centering
\tablestyle{7pt}{1.15}
\begin{tabular}{lccc}
\toprule
model & LN & DyT & change \\
\midrule
ViT-B & 82.3\% & 82.5\% & \better{0.2\%} \\
ViT-L & 83.1\% & 83.6\% & \better{0.5\%} \\
ConvNeXt-B & 83.7\% & 83.7\% & - \\
ConvNeXt-L & 84.3\% & 84.4\% & \better{0.1\%} \\
\midrule
  \end{tabular}
\caption{\textbf{Supervised classification accuracy on ImageNet-1K.} DyT achieves better or similar performance than LN across both architectures and model sizes.}
\label{table:classification}
\end{table}

\paragraph{Self-supervised learning in vision.}
We benchmark with two popular visual self-supervised learning methods: masked autoencoders (MAE) \citep{he2022masked} and DINO \citep{caron2021emerging}. Both by default use Vision Transformers as the backbones, but have different training objectives: MAE is trained with a reconstruction loss, and DINO uses a joint-embedding loss \citep{lecun2022path}.
Following the standard self-supervised learning protocol, we first pretrain models on ImageNet-1K without using any labels and then test the pretrained models by attaching a classification layer and fine-tuning them with labels.
The fine-tuning results are presented in Table~\ref{table:self_supervised_learning}. DyT consistently performs on par with LN in self-supervised learning tasks.

\begin{table}[h]
\centering
\tablestyle{7pt}{1.15}
\begin{tabular}{lccc}
\toprule
model & LN & DyT & change \\
\midrule
MAE ViT-B & 83.2\% & 83.2\% & - \\
MAE ViT-L & 85.5\% & 85.4\% & \worse{0.1\%} \\
DINO ViT-B (patch size 16) & 83.2\% & 83.4\% & \better{0.2\%} \\
DINO ViT-B (patch size 8) & 84.1\% & 84.5\% & \better{0.4\%} \\
\midrule
\end{tabular}
\caption{\textbf{Self-supervised learning accuracy on ImageNet-1K.} DyT performs on par with LN across different pretraining methods and model sizes in self-supervised learning tasks. }
\label{table:self_supervised_learning}
\end{table}

\paragraph{Diffusion models.} We train three Diffusion Transformer (DiT) models \citep{peebles2023scalable} of sizes B, L and XL on ImageNet-1K \citep{deng2009imagenet}. The patch size is 4, 4, and 2, respectively. Note that in DiT, the LN layers' affine parameters are used for class conditioning in DiT, and we keep them that way in our DyT experiments, only replacing the normalizing transformation with the $\tanh(\alpha \vx)$ function. After training, we evaluate the Fréchet Inception Distance (FID) scores using the standard ImageNet ``reference batch'', as presented in Table~\ref{table:diffusion}.
DyT achieves comparable or improved FID over LN.

\begin{table}[h!]
\centering
\tablestyle{7pt}{1.15}
\begin{tabular}{lccc}
\toprule
model & LN & DyT & change  \\
\midrule
DiT-B & 64.9 & 63.9 & \betterinv{1.0} \\
DiT-L & 45.9 & 45.7 & \betterinv{0.2} \\
DiT-XL & 19.9 & 20.8 & \worseinv{0.9} \\
\midrule
\end{tabular}
\caption{\textbf{Image generation quality (FID, lower is better) on ImageNet.} DyT achieves comparable or superior FID scores to LN across various DiT model sizes.}
\label{table:diffusion}
\end{table}

\begin{figure*}[t]
\vspace{-0.5in}
\centering
\begin{minipage}{0.49\textwidth}
\hspace*{-0.25cm}
\includegraphics[width=\textwidth]{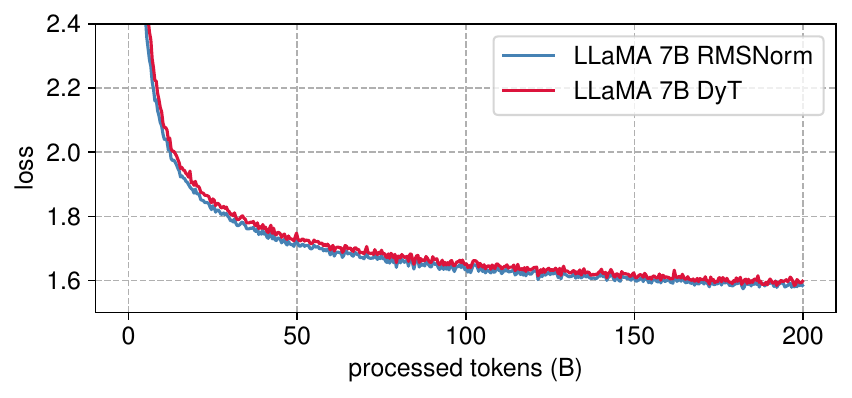}
\end{minipage}
\hfill
\begin{minipage}{0.49\textwidth}
\hspace*{-0.6cm}
  \includegraphics[width=\textwidth]{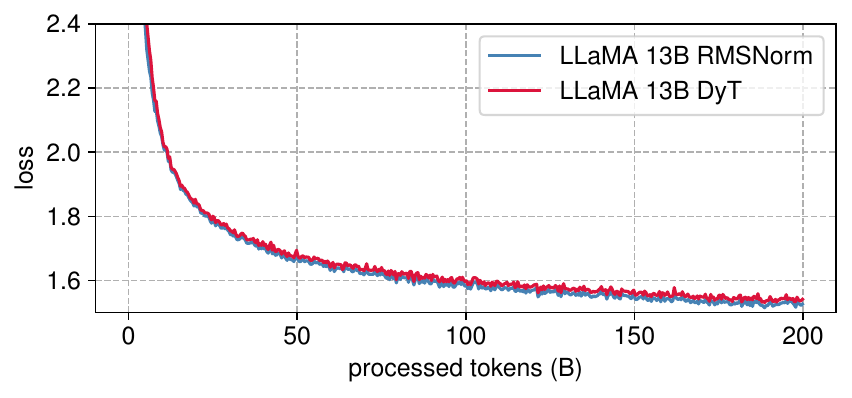}
\end{minipage}
\vfill
\begin{minipage}{0.49\textwidth}
\vspace*{-0.1cm}
\hspace*{-0.25cm}
\includegraphics[width=\textwidth]{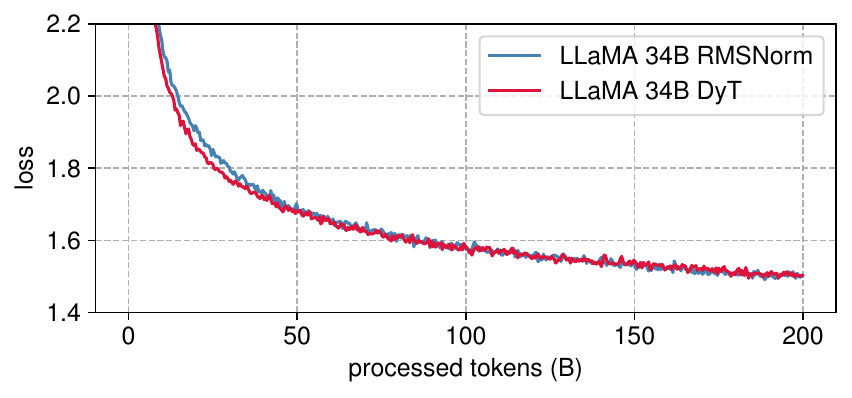}
\end{minipage}
\hfill
\begin{minipage}{0.49\textwidth}
\vspace*{-0.1cm}
\hspace*{-0.6cm}
\includegraphics[width=\textwidth]{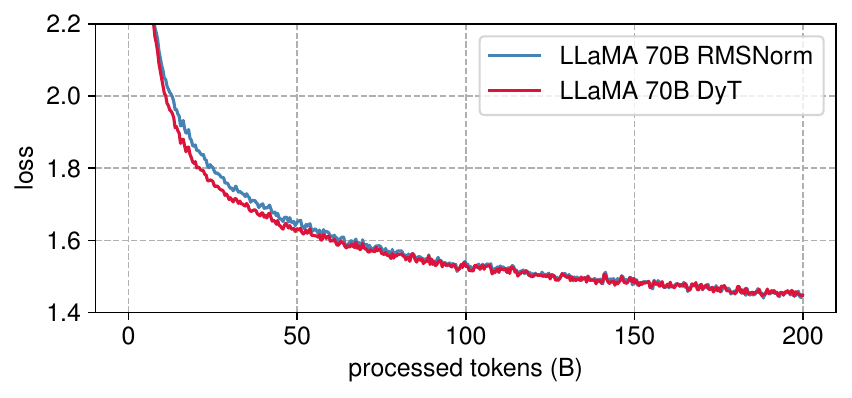}
\end{minipage}
\caption{\textbf{LLaMA pretraining loss.} The loss curves of DyT and RMSNorm models are closely aligned across model sizes.}
\label{figure:llama_curve}
\end{figure*}

\vskip -0.3in
\paragraph{Large Language Models.}
We pretrain LLaMA 7B, 13B, 34B, and 70B models~\citep{touvron2023llama, touvron2023llama2, dubey2024llama} to assess DyT performance relative to RMSNorm \citep{zhang2019root}, the default normalization layer used in LLaMA.
The models are trained on The Pile dataset~\citep{pile} with 200B tokens, following the original recipe outlined in LLaMA~\citep{touvron2023llama2}.
On LLaMA with DyT, we add a learnable scalar parameter after the initial embedding layer, and adjust the initial value of $\alpha$, as detailed in Section~\ref{section:alpha_init}.
We report the loss value after training and also follow OpenLLaMA~\citep{openlm2023openllama} to benchmark the models on 15 zero-shot tasks from \texttt{lm-eval} \citep{eval-harness}. As shown in Table~\ref{table:llama}, DyT performs on par with RMSNorm across all four model sizes. Figure~\ref{figure:llama_curve} illustrates the loss curves, demonstrating similar trends across all model sizes, with training losses closely aligned throughout training.

\begin{table}[h]
\centering
\tablestyle{10pt}{1.15}
\begin{tabular}{lccc}
\toprule
score / loss & RMSNorm & DyT &  change \\
\midrule
LLaMA 7B & 0.513 / 1.59 &  0.513 / 1.60 & - / \worseinv{0.01} \\
LLaMA 13B & 0.529 / 1.53 &  0.529 / 1.54 & - / \worseinv{0.01} \\
LLaMA 34B & 0.536 / 1.50 &  0.536 / 1.50 & - / - \\
LLaMA 70B & 0.549 / 1.45 &  0.549 / 1.45 & - / - \\
\midrule
\end{tabular}
\caption{\textbf{Language models' training loss and average performance with 15 zero-shot \texttt{lm-eval} tasks.} 
DyT achieves a comparable zero-shot performance and training loss to RMSNorm.}
\label{table:llama}
\end{table}

\vskip -0.1in
\paragraph{Self-supervised learning in speech.}
We pretrain two wav2vec 2.0 Transformer models \citep{baevski2020wav2vec} on the LibriSpeech dataset \citep{panayotov2015librispeech}.
We report the final validation loss in Table \ref{table:wav2vec2}. We observe that DyT performs comparably to LN in both model sizes.

\begin{table}[h!]
\centering
\tablestyle{7pt}{1.15}
\begin{tabular}{lccc}
\toprule
model & LN & DyT & change   \\
\midrule
wav2vec 2.0 Base &  1.95 & 1.95 & - \\
wav2vec 2.0 Large & 1.92 & 1.91 & \betterinv{0.01}  \\
\midrule
\end{tabular}
\caption{\textbf{Speech pretraining validation loss on LibriSpeech.} DyT performs comparably to LN for both wav2vec 2.0 models.}
\label{table:wav2vec2}
\end{table}

\vskip -0.1in
\paragraph{DNA sequence modeling.}
On the long-range DNA sequence modeling task, we pretrain the HyenaDNA model \citep{nguyen2024hyenadna} and the Caduceus model \citep{schiff2024caduceus}. The pretraining uses the human reference genome data from \citep{grch382013p13}, and the evaluation is on GenomicBenchmarks \citep{grevsova2023genomic}. The results are presented in Table \ref{table:dna_sequence}. DyT maintains performance comparable to LN for this task.

\begin{table}[h!]
\centering
\tablestyle{7pt}{1.15}
\begin{tabular}{lccc}
\toprule
model &LN & DyT & change \\
\midrule
HyenaDNA \citep{nguyen2024hyenadna} & 85.2\% & 85.2\% & - \\
Caduceus \citep{schiff2024caduceus} & 86.9\% & 86.9\% & - \\
\midrule
  \end{tabular}
  \caption{\textbf{DNA classification accuracy on GenomicBenchmarks}, averaged over each dataset in GenomicBenchmarks. DyT achieves comparable performance to LN.}
  \label{table:dna_sequence}
  \vspace{-0.2in}
\end{table}

\section{Analysis}

\vspace{-0.05in}
In this section, we begin with ablations on the effects of the tanh function and the learnable scalar $\alpha$. We then analyze the values of $\alpha$ throughout and after training. Lastly, we present comparisons with previous methods that aim to remove normalization layers.

\subsection{Ablations of tanh and $\alpha$}

To further investigate the role of tanh and $\alpha$ in DyT, we conduct experiments to evaluate the model’s performance when these components are altered or removed.

\paragraph{Replacing and removing tanh.} We replace tanh in DyT layers with alternative squashing functions, specifically hardtanh and sigmoid (Figure~\ref{figure:hardtanh}), while keeping the learnable scaler $\alpha$ intact. Furthermore, we assess the impact of completely removing tanh by replacing it with the identity function while still retaining $\alpha$.
As shown in Table \ref{tab:function_ablation}, the squashing function is essential for stable training. Using the identity function leads to unstable training and divergence, whereas squashing functions enable stable training. Among the squashing functions, tanh performs the best. This is possibly due to its smoothness and zero-centered properties.

\begin{table}[h]

\centering
\tablestyle{10pt}{1.15}
\begin{tabular}{lcccccc}
\toprule
model & identity & tanh & hardtanh & sigmoid \\
\midrule
ViT-S  & 58.5\% $\rightarrow$ failed & \textbf{80.3\%} & 79.9\% & 79.6\% \\
ViT-B  & 61.0\% $\rightarrow$ failed & \textbf{82.5\%} & 82.2\% & 81.6\% \\
\midrule
\end{tabular}
\caption{\textbf{ImageNet-1K classification accuracy with different squashing functions.} All experiments follow the same training recipe as the original LN-based models. Squashing functions play a crucial role in preventing divergence, with tanh achieving the highest performance among the three functions. ``$\rightarrow$ failed'' indicates that training diverged after some progress, with the preceding number representing the highest accuracy reached before divergence.}
\label{tab:function_ablation}
\end{table}

\paragraph{Removing $\alpha$.} Next, we evaluate the impact of removing the learnable $\alpha$ while retaining the squashing functions (tanh, hardtanh, and sigmoid).
As shown in Table~\ref{tab:alpha_ablation}, removing $\alpha$ results in performance degradation across all squashing functions, highlighting the critical role of $\alpha$ in overall model performance.

\begin{table}[h]
\centering
\tablestyle{7pt}{1.15}
\begin{tabular}{lcccccc}
\toprule
model & tanh & hardtanh & sigmoid \\
\midrule
without $\alpha$ &  81.1\% & 80.7\% & 80.7\% \\
with $\alpha$  & \textbf{82.5\%} & \textbf{82.2\%} & \textbf{81.6\%} \\
\midrule
\end{tabular}
\caption{\textbf{ImageNet-1K classification accuracy with ViT-B.} All experiments follow the same training recipe as the original LN-based models. The learnable $\alpha$ is essential for enhancing model performance.}
\label{tab:alpha_ablation}
\end{table}

\subsection{Values of $\alpha$}

\begin{figure}[t]
\vspace{-0.3in}
\centering
\begin{minipage}{0.39\textwidth}
\centering
\vspace{0.05cm}
\includegraphics[width=0.96\linewidth]{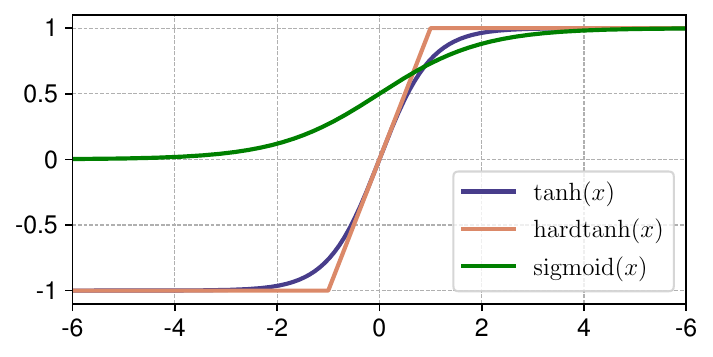}
\vspace{0.3cm}
\caption{Curves of three squashing functions: tanh, hardtanh, and sigmoid. All three functions squash inputs into a bounded range, but $\tanh(x)$ achieves the best performance when used in DyT layers. We suspect it is due to its smoothness and zero-centered properties.}

\label{figure:hardtanh}
\end{minipage}
\hfill
\begin{minipage}{0.58\textwidth}
\centering
\begin{minipage}{0.48\textwidth}
\includegraphics[width=\linewidth]{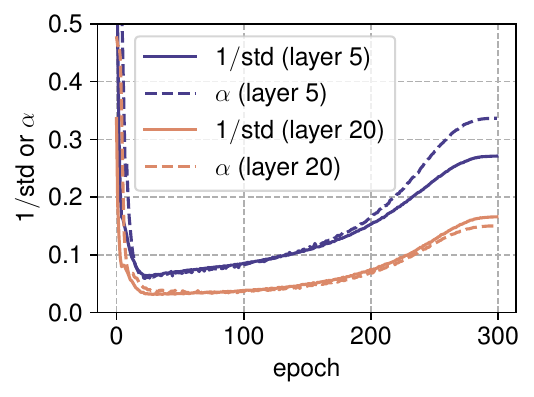}
\end{minipage}
\hfill
\begin{minipage}{0.48\textwidth}
\includegraphics[width=\linewidth]{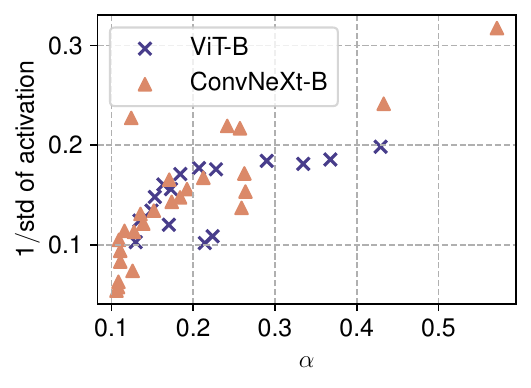}
\end{minipage}
\caption{\emph{Left:} For two selected DyT layers from the ViT-B model, we track $\alpha$ and the inverse of the standard deviation ($1/\mathrm{std}$) of activations at the end of each epoch, observing that they evolve together during training. \emph{Right:} We plot the final $\alpha$ values of two trained models, ViT-B and ConvNeXt-B, against the $1/\mathrm{std}$ of the input activations, demonstrating a strong correlation between the two values.}
\label{figure:all_three}
\end{minipage}
\end{figure}

\paragraph{During training.} Our analysis reveals that the $\alpha$ closely tracks the $1/\mathrm{std}$ of activations throughout training. As illustrated in the left panel of Figure \ref{figure:all_three}, $\alpha$ first decrease and then increase during training, but always fluctuate consistently with the standard deviation of input activations. This supports the important role of $\alpha$ in maintaining activations within a suitable range, which leads to stable and effective training.

\paragraph{After training.}
Our further analysis of the final values of $\alpha$ in trained networks reveals a strong correlation with the $1/\mathrm{std}$ of the input activations.
As shown on the right panel of Figure~\ref{figure:all_three}, higher $1/\mathrm{std}$ values generally correspond to larger $\alpha$ values, and vice versa. Additionally, we observe that deeper layers tend to have activations with larger standard deviations. This trend aligns with characteristics of deep residual networks, as shown in \citet{brock2021characterizing} for ConvNets, and \citet{sun2025cursedepthlargelanguage} for Transformers.

Both analyses suggest that $\alpha$ functions partially as a normalization mechanism by learning values approximating $1/\mathrm{std}$ of the input activations. Unlike LN, which normalizes the activations per token, $\alpha$ normalizes the entire input activations collectively. Consequently, $\alpha$ alone cannot suppress extreme values in a non-linear fashion.

\subsection{Comparison with Other Methods}

To further assess DyT's effectiveness, we compare it with other methods that also enable training Transformers without normalization layers. These methods can be broadly categorized into initialization-based and weight-normalization-based methods.
We consider two popular initialization-based methods, Fixup~\citep{zhang2019fixup, huang2020improving} and SkipInit~\citep{de2020batch, bachlechner2021rezero}.
Both methods aim to mitigate training instabilities by adjusting the initial parameter values to prevent large gradients and activations at the start of training, thereby enabling stable learning without normalization layers.
In contrast, weight-normalization-based methods impose constraints on network weights throughout training to maintain stable learning dynamics in the absence of normalization layers. We include one such method, $\sigma$Reparam~\citep{zhai2023stabilizing}, which controls the spectral norm of the weights to promote stable learning.

\begin{table}[h]
\centering
\tablestyle{2pt}{1.15} 
\begin{tabular}{lx{40}x{40}x{40}x{40}x{40}x{40}}
\toprule
model & LN & Fixup & SkipInit & $\sigma$Reparam & DyT \\ 
\midrule
ViT-B & 82.3\%  & 77.2\% & 74.1\% & 82.5\% & \textbf{82.8\%} \\
ViT-L & 83.1\%  & 78.1\% & 75.6\% & 83.0\% & \textbf{83.6\%}  \\
\midrule
MAE ViT-B & 83.2\%  & 73.7\% & 73.1\% & 83.2\% & \textbf{83.7\%} \\
MAE ViT-L & 85.5\%   & 74.1\% & 74.0\% & 85.4\% & \textbf{85.8\%} \\
\midrule
  \end{tabular}
\caption{\textbf{Classification accuracy on ImageNet-1K.}
DyT consistently achieves superior performance over other methods.}
\label{table:compare}
\vspace{-0.8em}
\end{table}

Table~\ref{table:compare} summarizes the results of two ViT-based tasks. We closely follow the original protocols outlined in their respective papers. However, we find that both initialization-based methods, Fixup and SkipInit, require significantly lower learning rates to prevent training divergence. To ensure a fair comparison, we conduct a simple learning rate search for all methods, including DyT. This produces results that differ from those reported in Section~\ref{section:experiments}, where no hyperparameter is tuned. Overall, the results show that DyT consistently outperforms all other tested methods across different configurations.

\section{Initialization of $\alpha$}
\label{section:alpha_init}

We find that tuning the initialization of $\alpha$ (denoted $\alpha_0$) rarely leads to significant performance improvements.
The only exception is LLM training, where careful tuning of $\alpha_0$ yields noticeable performance gains.
In this section, we detail our findings on the impact of $\alpha$ initialization.

\subsection{Initialization of $\alpha$ for Non-LLM Models}

\begin{figure*}[t]
\vskip -0.3in
\centering
\includegraphics[width=\textwidth]{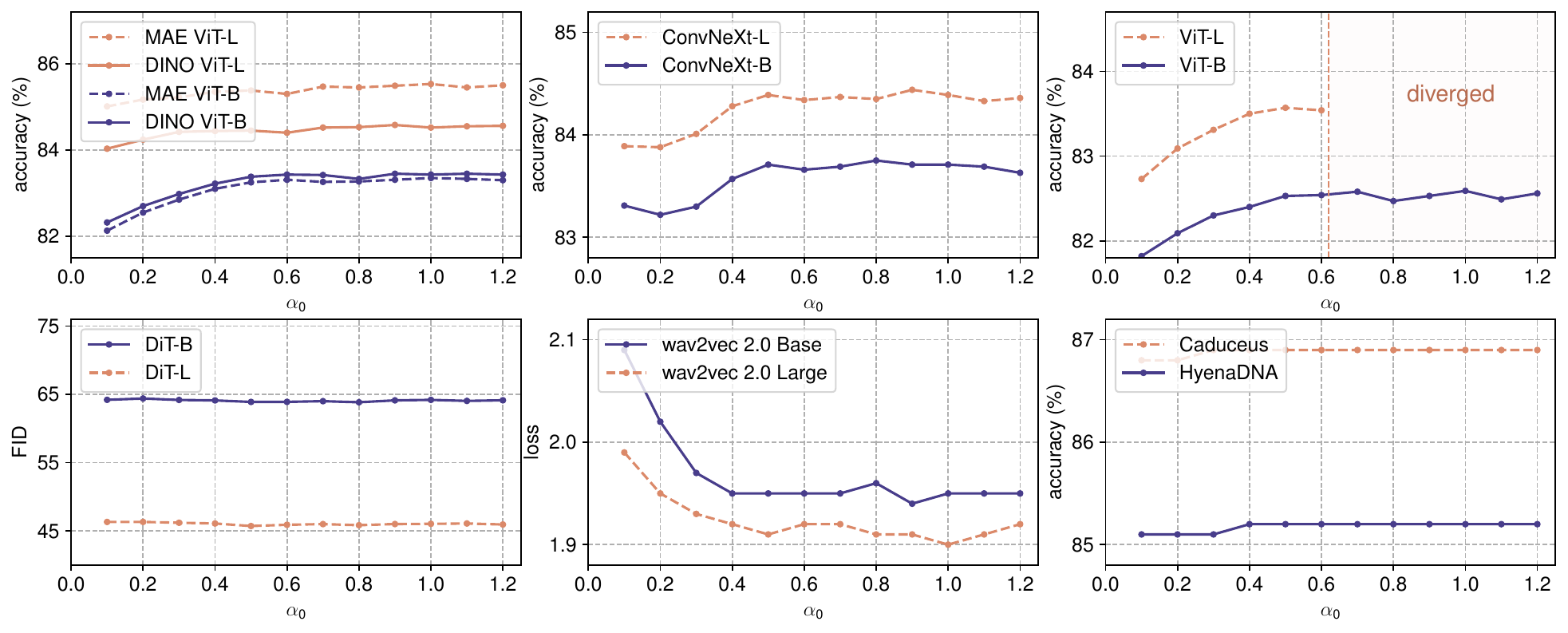}
\caption{\textbf{Performance of different tasks across different $\alpha_0$ values.} We benchmark the performance of all non-LLM tasks used in Section \ref{section:experiments} with different initial values of $\alpha$. Performance remains stable across a wide range of $\alpha_0$ values. The only exception is that supervised ViT-L models (top right panel) will diverge for $\alpha_0$ values larger than 0.6.}\label{figure:accuracy_vs_alpha}
\vskip -0.15in
\end{figure*}

\paragraph{Non-LLM models are relatively insensitive to $\alpha_0$.}
Figure~\ref{figure:accuracy_vs_alpha} shows the effect of varying $\alpha_0$ on validation performance across different tasks. All experiments follow the original setup and hyperparameters of their respective recipe.
We observe that performance remains stable across a wide range of $\alpha_0$ values, with values between 0.5 and 1.2 generally yielding good results. We observe that adjusting $\alpha_0$ typically affects only the early stages of the training curves.
The main exception is supervised ViT-L experiments, where training becomes unstable and diverges when $\alpha_0$ exceeds 0.6. In such cases, reducing the learning rate restores stability, as detailed below.

\begin{figure*}[b] 
\centering \includegraphics[width=0.98\textwidth]{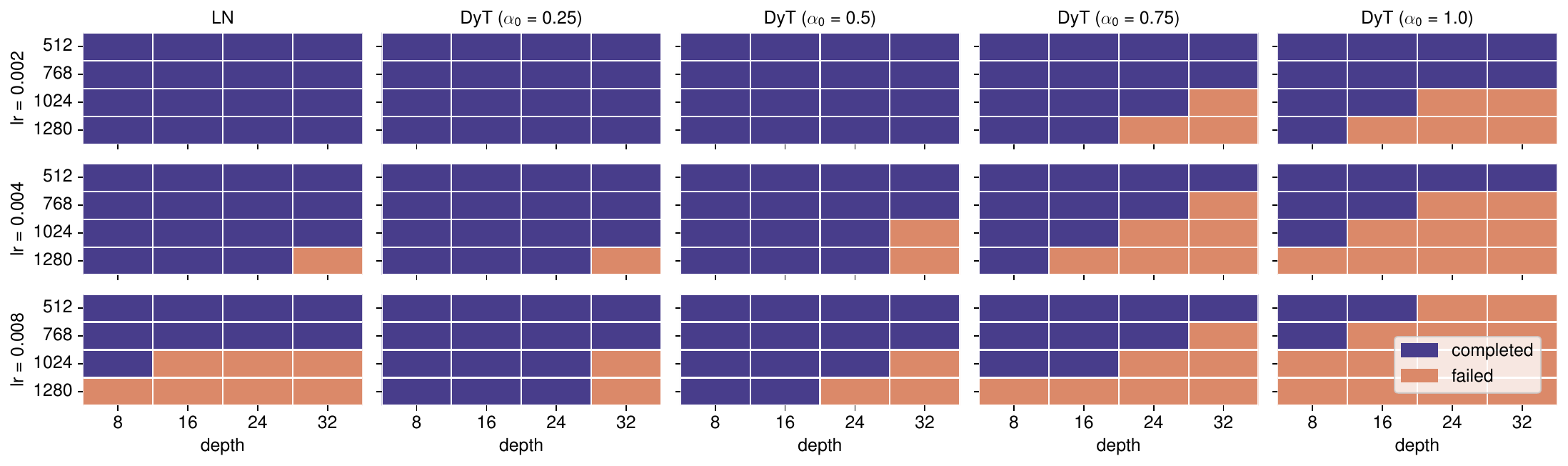} \caption{\textbf{Stability across varying $\alpha_0$ values, learning rates, and model sizes.} We train supervised ViT models on the ImageNet-1K dataset and observe that larger models are more prone to instability for both LN and DyT models. Lowering the learning rate or reducing $\alpha_0$ enhances stability. LN shows similar stability to DyT with $\alpha_0 = 0.5$.}

\label{figure:alpha_lr}
\end{figure*} 

\paragraph{Smaller $\alpha_0$ results in more stable training.} Building on previous observations, we further analyze the factors contributing to training instability. Our findings suggest that increasing either the model size or the learning rate requires lowering $\alpha_0$ to ensure stable training. Conversely, a higher $\alpha_0$ requires a lower learning rate to mitigate training instability.
Figure~\ref{figure:alpha_lr} shows the ablation of the training stability of supervised ViT with ImageNet-1K dataset. We vary learning rates, model sizes, and $\alpha_0$ values. Training a larger model is more prone to failure, requiring smaller $\alpha_0$ values or learning rates for stable training. A similar instability pattern is also observed in LN-based models under comparable conditions, and setting $\alpha_0 = 0.5$ results in a stability pattern similar to that of LN.

\paragraph{Setting $\alpha_0 = 0.5$ as the  default.} Based on our findings, we set $\alpha_0 = 0.5$ as the default value for all non-LLM models.
This setting provides training stability comparable to LN while maintaining strong performance.

\subsection{Initialization of $\alpha$ for LLMs}

\paragraph{Tuning $\alpha_0$ enhances LLM performance.} As discussed earlier, the default setting of $\alpha_0 = 0.5$ generally performs well across most tasks. However, we find tuning $\alpha_0$ can substantially improve LLM performance. We tune $\alpha_0$ across LLaMA models by pretraining each on 30B tokens and comparing their training losses. Table~\ref{table:alpha_init_llama} summarizes the tuned $\alpha_0$ values for each model. Two key findings emerge:
\begin{enumerate}
\item \textbf{Larger models require smaller $\alpha_0$ values.} Once the optimal $\alpha_0$ is determined for smaller models, the search space for larger models can be reduced accordingly.
\item \textbf{Higher $\alpha_0$ values for attention blocks improve performance.} We find that initializing $\alpha$ with higher values for DyT layers in attention blocks and lower values for DyT layers in other locations (i.e., within FFN blocks or before the final linear projection) improves performance. 
\end{enumerate}

\begin{table}[h]
\centering
\tablestyle{3pt}{1.15}
\begin{tabular}{y{50}x{30}x{30}x{60}}
\toprule
\multirow{2}{*}{model} & \multirow{2}{*}{width} & \multirow{2}{*}{depth} & optimal $\alpha_0$ \\
& &  & (attention/other) \\
\midrule
LLaMA 7B & 4096 & 32 & 0.8/0.2  \\
LLaMA 13B & 5120 & 40 & 0.6/0.15 \\
LLaMA 34B & 8196 & 48 & 0.2/0.05 \\
LLaMA 70B & 8196 & 80 &  0.2/0.05 \\
\midrule
\end{tabular}
\caption{\textbf{Optimal $\alpha_0$ for different LLaMA models.} Larger models require smaller $\alpha_0$ values. We find it is important to initialize $\alpha$ differently in (1) attention blocks (``attention''), versus (2) the FFN blocks, and the final DyT layer before outputs (``other''). $\alpha_0$ in attention blocks require larger values.}
\label{table:alpha_init_llama}
\end{table}

\begin{figure}[h]
\centering
\begin{minipage}{0.49\textwidth}
\includegraphics[width=1.04\linewidth]{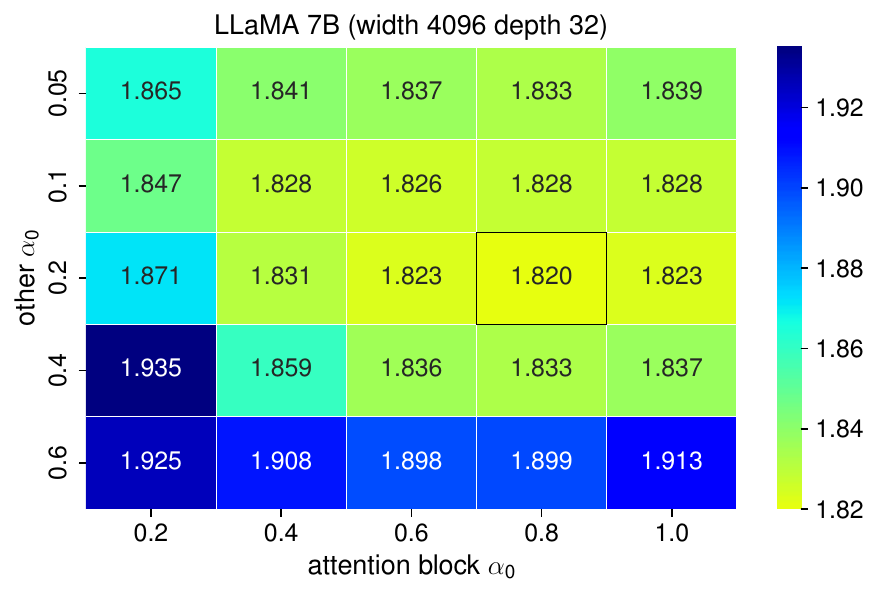}
\end{minipage}
\hfill
\begin{minipage}{0.49\textwidth}
\includegraphics[width=1.04\linewidth]{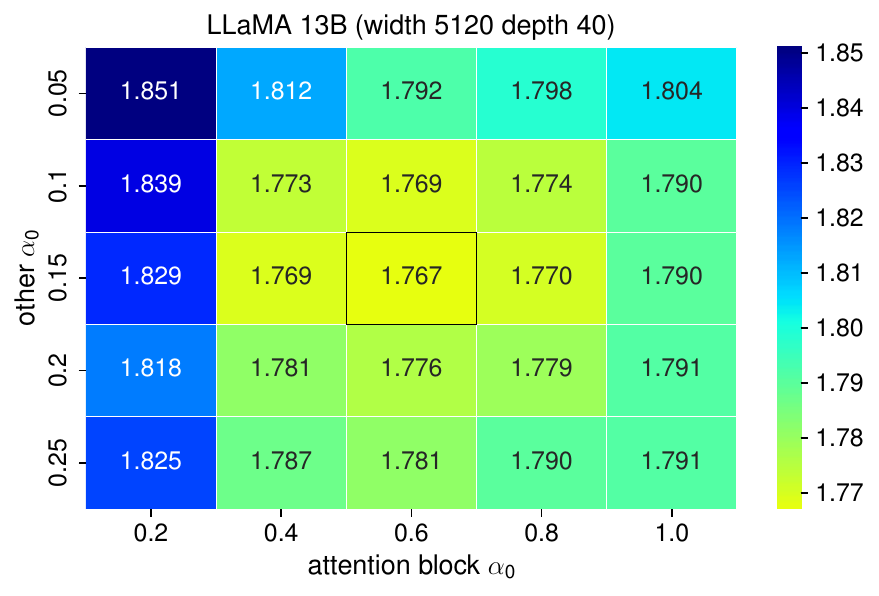}
\end{minipage}
\hfill
\caption{\textbf{Heatmaps of loss values at 30B tokens for different $\alpha_0$ settings.} Both LLaMA models benefit from increased $\alpha_0$ in attention blocks.} 
\label{figure:heat}
\end{figure}

To further illustrate the impact of $\alpha_0$ tuning, Figure~\ref{figure:heat} presents heatmaps of loss values of two LLaMA models. Both models benefit from higher $\alpha_0$ in attention blocks, leading to reduced training loss. 

\paragraph{Model width primarily determines $\alpha_0$ selection.} We also investigate the influence of model width and depth on the optimal $\alpha_0$. We find that the model width is critical in determining the optimal $\alpha_0$, while model depth has minimal influence. Table~\ref{table:width_depth} shows the optimal $\alpha_0$ values across different widths and depths, showing that wider networks benefit from smaller $\alpha_0$ values for optimal performance. On the other hand, model depth has negligible impact on the choice of $\alpha_0$.

\begin{table}[t]
\vskip -0.1in
\centering
\tablestyle{7pt}{1.15}
\begin{tabular}{ccccccccccc}
\toprule
 width / depth & 8 & 16 & 32 & 64 \\
\midrule
1024 & 1.0/1.0 & 1.0/1.0 & 1.0/1.0 & 1.0/1.0 \\
2048 & 1.0/0.5 & 1.0/0.5 & 1.0/0.5 & 1.0/0.5 \\
4096 & 0.8/0.2 & 0.8/0.2 & 0.8/0.2 & 0.8/0.2 \\
8192 & 0.2/0.05 & 0.2/0.05 & 0.2/0.05 & 0.2/0.05 \\
\midrule
\end{tabular}
\caption{\textbf{Optimal $\alpha_0$ (attention / other) across model widths and depths in LLaMA training.} Model width significantly impacts the choice of $\alpha_0$, with wider networks requiring smaller values. In contrast, model depth has negligible influence.}
\label{table:width_depth}
\end{table}

As can be seen in Table~\ref{table:width_depth}, the wider the network, the more uneven initialization for ``attention'' and ``other'' is needed. We hypothesize that the sensitivity of LLM's $\alpha$ initialization is related to their excessively large widths compared to other models.

\section{Related Work}

\paragraph{Mechanisms of Normalization layers.}
There has been a rich line of work investigating normalization layers' role in enhancing model performance through various mechanisms.
These include stabilizing gradient flow during training \citep{balduzzi2017shattered, daneshmand2020batch, lubana2021beyond}, reducing sensitivity to weight initialization \citep{zhang2019fixup, de2020batch, shao2020normalization}, moderating outlier eigenvalues \citep{bjorck2018understanding,karakida2019normalization}, auto-tuning learning rates \citep{arora2018theoretical,tanaka2021noether}, and smoothing the loss landscape for more stable optimization \citep{santurkar2018does}.
These earlier works focused on studying batch normalization. Recent studies \citep{lyu2022understanding, dai2024crucial, mueller2024normalization} further highlight the connection between normalization layers and sharpness reduction, which contributes to better generalization.

\paragraph{Normalization in Transformers.} With the rise of Transformer \citep{vaswani2017attention}, research has increasingly focused on layer normalization \citep{ba2016layer}, which has proven particularly effective for sequential data in natural language tasks \citep{nguyen2019transformers, xu2019understanding, xiong2020layer}. 
Recent work \citep{ni2024nonlinearity} reveals that layer normalization introduces strong non-linearity, enhancing the model's representational capacity. Additionally, studies \citep{loshchilov2024ngpt, li2024mix} demonstrate that modifying the location of normalization layers within Transformers can improve convergence properties.

\paragraph{Removing normalization.}
Many studies have explored how to train deep models without normalization layers.  \citet{klambauer2017self} introduce an alternative activation function that enables self-normalizing behavior, eliminating the need for explicit normalization.
Other works~\citep{zhang2019fixup, de2020batch, bachlechner2021rezero} propose specialized initialization schemes to stabilize training in the absence of normalization.
The pioneering work by \citet{brock2021characterizing, brock2021high} show that high-performing ResNets can be trained without normalization \citep{smith2023convnets} through combination of initialization techniques \citep{de2020batch},  weight normalization \citep{salimans2016weight, huang2017centered, qiao2019micro}, and adaptive gradient clipping~\citep{brock2021high}. Additionally, their training strategy incorporates extensive data augmentation \citep{cubuk2020randaugment} and regularization \citep{srivastava2014dropout, huang2016deep}. The studies above are based on various ConvNet models.

In Transformer architectures, \citet{he2023simplifying} explore modifications to Transformer blocks that reduce reliance on normalization layers and skip connections. \citet{jha2024aero} introduce AERO, a Softmax-only LLM that improves inference efficiency and privacy with minimal performance loss. Alternatively, \citet{heimersheim2024you} propose a method to gradually remove LN from pretrained networks by fine-tuning the model after removing each normalization layer.
Unlike previous approaches, DyT requires minimal modifications to both the architecture and the training recipe. Despite its simplicity, DyT achieves stable training and comparable performance.

\vspace{-0.2cm}
\section{Limitations}

Our experiments focus on networks using LN or RMSNorm because of their popularity in Transformers and other modern architectures. Preliminary experiments (see Appendix \ref{section:batch_normalization}) indicate that DyT struggles to replace BN directly in classic networks like ResNets. It remains to be studied in more depth whether and how DyT can adapt to models with other types of normalization layers.

Furthermore, although DyT is conceptually and computationally simpler, we find that DyT offers no speedup over models with normalization layers when properly compiled/optimized (see Appendix \ref{section: efficiency}). Its computational benefits across different hardware platforms or deployment environments remain uncertain.

\vspace{-0.2cm}
\section{Conclusion}

In this work, we demonstrate modern neural networks, in particular Transformers, can be trained without normalization layers. This is done through Dynamic Tanh (DyT), a simple replacement for traditional normalization layers. It adjusts the input activation range via a learnable scaling factor $\alpha$ and then squashes the extreme values through an $S$-shaped tanh function. Although a simpler function, it effectively captures the behavior of normalization layers. Under various settings, models with DyT match or exceed the performance of their normalized counterparts. The findings challenge the conventional understanding of the necessity of normalization layers in training modern neural networks. Our study also contributes to understanding the mechanisms of normalization layers, one of the most fundamental building blocks in deep neural networks.

\bibliographystyle{assets/plainnat}
\bibliography{paper}

\clearpage
\newpage
\beginappendix

\appendix

\section{Experimental Settings}
\label{section:reproduce}

\paragraph{Supervised image classification.} For all supervised classification experiments on ImageNet-1K, we follow the training recipes from ConvNeXt \citep{convnext}.
For ConvNeXt-B and ConvNeXt-L, we use the original hyperparameters without modification.
ViT-B and ViT-L models use the same hyperparameters as ConvNeXt-B, except that for ViT-L, the beta parameters for AdamW are set to (0.9, 0.95), and the stochastic depth rates are set to 0.1 for ViT-B and 0.4 for ViT-L. 

\paragraph{Diffusion models.} We use the official implementation~\citep{dit} for training all DiT models. We find that the default learning rate is suboptimal for the models considered in this paper. To address this, we conduct a simple learning rate search with the LN models and apply the tuned learning rates directly to the DyT models. We also observe that the zero initialization negatively affects the performance of DyT models. Therefore, we retain the zero initialization for LN models but remove the zero initialization for DyT models.

\paragraph{Large Language Models.} In our implementation of LLaMA models~\citep{touvron2023llama, touvron2023llama2, dubey2024llama} with DyT, we introduce an additional learnable scalar parameter immediately after the embedding layer, before any Transformer blocks. We initialize it to the square root of the model embedding dimension $\sqrt{d}$. Without this scaling scalar, we find that the magnitudes of model activations at the beginning of training are too small, and the training struggles to progress. The issue is mitigated by incorporating a learnable scalar, and the model can converge normally. This addition of a scalar is similar to the original Transformer~\citep{vaswani2017attention} design, which uses a fixed scalar of the same value at the same position.

We train all our LLaMA models on the Pile dataset~\citep{pile}. We use the codebase from \texttt{FMS-FSDP} \citep{fms-fsdp}, which provides a default training recipe for the 7B model that closely follows the LLaMA 2 paper~\citep{touvron2023llama2}. We maintain the learning rate at the default 3e-4 for 7B and 13B and 1.5e-4 for 34B and 70B, in line with LLaMA 2.
The batch size is set to 4M tokens
and each model is trained on a total of 200B tokens.

For evaluation, we test the pretrained models on 15 zero-shot commonsense reasoning tasks from \texttt{lm-eval} \citep{eval-harness}: \texttt{anli\_r1}, \texttt{anli\_r2}, \texttt{anli\_r3}, \texttt{arc\_challenge}, \texttt{arc\_easy}, \texttt{boolq}, \texttt{hellaswag}, \texttt{openbookqa}, \texttt{piqa}, \texttt{record}, \texttt{rte}, \texttt{truthfulqa\_mc1}, \texttt{truthfulqa\_mc2}, \texttt{wic}, and \texttt{winogrande}. The selection closely follows that of OpenLLaMA~\citep{openlm2023openllama}. We report the average performance across all tasks.

\paragraph{Self-supervised learning in speech.} For both wav2vec 2.0 models, we retain the first group normalization layer from the original architecture, as it functions primarily as data normalization to handle the unnormalized input data.
We use the official implementation \citep{wav2vec2} without modifying hyperparameters for both the Base and Large models. We report the final validation loss.

\paragraph{Other tasks.} For all other tasks, MAE \citep{he2022masked}, DINO \citep{caron2021emerging}, HyenaDNA \citep{nguyen2024hyenadna} and Caduceus \citep{schiff2024caduceus}, we directly use the publicly released code \citep{mae, dino, hyena, caduceus}, without hyperparameter tuning, for both models with LN and DyT.

\section{Hyperparameters}
\label{section:tuning}

We present additional experiments to evaluate the impact of hyperparameter tuning, specifically focusing on the learning rate and initialization of $\alpha$ for all non-LLM models. 

\paragraph{Tuning learning rate.} Table~\ref{table:tuned_lr} summarizes performance comparisons between models trained with original versus tuned learning rates. Results indicate that tuning the learning rate provides only modest performance improvements for DyT models. This suggests that the original hyperparameters, initially optimized for LN models, are already well-suited for DyT models. This observation underscores the inherent similarity between the DyT and LN models.

\begin{table}[t]
\vskip -0.05in
\centering
\tablestyle{7pt}{1.15}
\begin{tabular}{lcccc}
\toprule
model & LN (original) & DyT (original) & LN (tuned) & DyT (tuned)  \\
\midrule
ViT-B & 82.3\% \scriptsize{(4e-3)} & {82.5\%} \scriptsize{(4e-3)} & - & {82.8\%} \scriptsize{(6e-3)} \\
ViT-L & 83.1\% \scriptsize{(4e-3)} & {83.6\%} \scriptsize{(4e-3)} & - & - \\
ConvNeXt-B & 83.7\% \scriptsize{(4e-3)} & 83.7\% \scriptsize{(4e-3)} & - & - \\
ConvNeXt-L & 84.3\% \scriptsize{(4e-3)} & {84.4\%} \scriptsize{(4e-3)} & - & - \\
\midrule
MAE ViT-B & 83.2\% \scriptsize{(2.4e-3)} & 83.2\% \scriptsize{(2.4e-3)} & - & 83.7\% \scriptsize{(3.2e-3)} \\
MAE ViT-L & {85.5\%} \scriptsize{(2.4e-3)} & 85.4\% \scriptsize{(2.4e-3)} & - & {85.8\%} \scriptsize{(3.2e-3)} \\
DINO ViT-B (patch size 16) & 83.2\% \scriptsize{(7.5e-4)} & {83.4\%} \scriptsize{(7.5e-4)} & 83.3\% \scriptsize{(1e-3)} & - \\
DINO ViT-B (patch size 8) & 84.1\% \scriptsize{(5e-4)} & {84.5\%} \scriptsize{(5e-4)} & - & - \\
\midrule
DiT-B & 64.9 \scriptsize{(4e-4)} & {63.9} \scriptsize{(4e-4)} & - & - \\
DiT-L & {45.9} \scriptsize{(4e-4)} & 45.7 \scriptsize{(4e-4)} & - & - \\
DiT-XL & {19.9} \scriptsize{(4e-4)}  & 20.8 \scriptsize{(4e-4)} & - & - \\
\midrule
wav2vec 2.0 Base & 1.95 \scriptsize{(5e-4)} & 1.95 \scriptsize{(5e-4)} & - & {1.94} \scriptsize{(6e-4)} \\
wav2vec 2.0 Large & 1.92 \scriptsize{(3e-4)} & {1.91} \scriptsize{(3e-4)} & - & - \\
\midrule
HyenaDNA & 85.2\% \scriptsize{(6e-4)} & 85.2\% \scriptsize{(6e-4)} & - & - \\
Caduceus & 86.9\% \scriptsize{(8e-3)} & 86.9\% \scriptsize{(8e-3)} &  - & - \\
\midrule
  \end{tabular}
\caption{\textbf{Performance comparison between original and tuned learning rates for LN and DyT models.} Results show that tuning learning rates provide only modest performance improvements for DyT models, suggesting that the default hyperparameters optimized for LN models are already well-suited for DyT models. Entries marked with ``-'' indicate no performance gain over the original learning rate. The values in parentheses represent the learning rate used. 
}
\label{table:tuned_lr}
\end{table}

\vspace{-1cm}
\paragraph{Tuning initial value of $\alpha$.} We also investigate the effects of optimizing $\alpha_0$ for DyT models, as presented in Table~\ref{table:tune_alpha}. Findings show only minor performance enhancements for select models when $\alpha_0$ is tuned, indicating that the default initial value ($\alpha_0 = 0.5$) generally achieves near-optimal performance.

\begin{table}[h]
\centering
\tablestyle{7pt}{1.15}
\begin{tabular}{lcccc}
\toprule
Model & LN  & DyT ($\alpha_0 = 0.5$) & DyT (tuned) \\
\midrule
ViT-B & 82.3\% & 82.5\% & 82.6\% \scriptsize{($\alpha_0 = 1.0$)} \\
ViT-L & 83.1\% & 83.6\% & - \\
ConvNeXt-B & 83.7\% & 83.7\% & - \\
ConvNeXt-L & 84.3\% & 84.4\% & - \\
\midrule
MAE ViT-B & 83.2\% & 83.2\% & 83.4\% \scriptsize{($\alpha_0 = 1.0$)} \\
MAE ViT-L & 85.5\% & 85.4\% & - \\
DINO ViT-B (patch 16) & 83.2\% & 83.4\% & - \\
DINO ViT-B (patch 8) & 84.1\% & 84.5\% & - \\
\midrule
DiT-B & 64.9 & 63.9 & - \\
DiT-L & 45.9 & 45.7 & - \\
DiT-XL & 19.9 & 20.8 & -  \\
\midrule
wav2vec 2.0 Base & 1.95 & 1.95 & - \\
wav2vec 2.0 Large & 1.92 & 1.91 & 1.90 \scriptsize{($\alpha_0 = 1.0$)} \\
\midrule
HyenaDNA & 85.2\% & 85.2\% & -  \\
Caduceus & 86.9\% & 86.9\% & - \\
\midrule
  \end{tabular}
 \caption{\textbf{Impact of tuning the $\alpha_0$ in DyT models.} Optimizing $\alpha_0$ from the default value ($\alpha_0 = 0.5$)  yields only minor performance gains for select DyT models, implying the default initialization already achieves near-optimal performance. Entries marked with ``-'' indicate no improvement over the default $\alpha_0$.
}
\label{table:tune_alpha}
\end{table}


\section{Efficiency of DyT}
\label{section: efficiency}

We benchmark the LLaMA 7B model with RMSNorm or DyT by measuring the total time required for 100 forward passes (inference) and 100 forward-backward passes (training) on a single sequence of 4096 tokens.
We first follow the officially recommended LLaMA setup and load the model from \citet{huggingface} without applying any performance optimizations.
Table~\ref{table:speed_latency} reports the time taken for RMSNorm and DyT layers, as well as for the entire model, when running on a Nvidia H100 GPU with BF16 precision. DyT layers reduce computation time compared to RMSNorm layers.

\begin{table}[h]
\vspace{0.1cm}
\centering
\tablestyle{6pt}{1.15}
\begin{tabular}{ccccc} 
\toprule
& \multicolumn{2}{c}{inference} & \multicolumn{2}{c}{training} \\
\cmidrule[0.5pt](lr){2-3} \cmidrule[0.5pt](lr){4-5}
LLaMA 7B & layer & model & layer & model \\ 
\midrule
RMSNorm & 2.1s & 14.1s & 8.3s & 42.6s \\
DyT & 1.0s & 13.0s & 4.8s & 39.1s \\
\midrule
reduction & \betterinv{52.4\%} & \betterinv{7.8\%} & \betterinv{42.2\%} & \betterinv{8.2\%}  \\
\midrule
\end{tabular}
\caption{\textbf{Inference and training latency (BF16 precision) for LLaMA 7B with RMSNorm or DyT.} DyT achieves a substantial reduction in both inference and training time. Results are measured without any extra performance optimizations.}
\label{table:speed_latency}
\vspace{0.1cm}
\end{table}

We also benchmark both models using \texttt{torch.compile}. Interestingly, compiling the entire LLaMA model increases latency for the \citet{huggingface} implementation, and compiling only the DyT or RMSNorm layers yields more efficient execution. Table~\ref{table:speed_compiled} shows that,
after compilation, the latency of the RMSNorm and DyT layers
becomes nearly identical.

\begin{table}[h]
\vspace{0.1cm}
\centering
\tablestyle{6pt}{1.15}
\begin{tabular}{ccccc} 
\toprule
& \multicolumn{2}{c}{inference} & \multicolumn{2}{c}{training} \\
\cmidrule[0.5pt](lr){2-3} \cmidrule[0.5pt](lr){4-5}
LLaMA 7B & layer & model & layer & model \\ 
\midrule
RMSNorm & 0.3s & 12.3s & 3.9s & 38.9s \\
DyT & 0.3s & 12.3s & 3.9s & 38.9s \\
\midrule
\end{tabular}
\caption{\textbf{Inference and training latency (BF16 precision) for a LLaMA 7B with compiled RMSNorm or DyT.} After compilation, the latency of RMSNorm and DyT layers are nearly identical.}
\label{table:speed_compiled}
\vspace{0.1cm}
\end{table}

An important distinction of DyT is that it is an element-wise operation and does not require a reduction operation within itself, compared to normalization layers. This could make it faster on hardware where reduction is a bottleneck. Additionally, even on conventional GPUs, DyT could offer opportunities for further optimization, e.g., fusing it with the preceding matrix multiplication layer from the last residual block.

\section{Replacing Batch Normalization with DyT}
\label{section:batch_normalization}

We investigate the potential of replacing BN with DyT in classic ConvNets such as ResNet-50~\citep{he2016deep} and VGG19~\citep{simonyan2014very}.
Both models are trained on the ImageNet-1K dataset~\citep{deng2009imagenet} using the training recipes provided by \texttt{torchvision}. The DyT models are trained using the same hyperparameters as their BN counterparts.

\begin{table}[h]
\centering
\tablestyle{7pt}{1.15}
\begin{tabular}{lcccccc}
\toprule
model & BN & DyT \\
\midrule
ResNet-50 & 76.2\% & 68.9\% \\
VGG19   & 72.7\% & 71.0\% \\
\midrule
\end{tabular}
\caption{\textbf{ImageNet-1K classification accuracy with BN and DyT.} Replacing BN with DyT in ResNet-50 and VGG19 results in a performance drop, indicating that DyT cannot fully substitute BN in these architectures.}
\label{table:bn_ablation}
\end{table}

The results are summarized in Table~\ref{table:bn_ablation}. Replacing BN with DyT led to a noticeable drop in classification accuracy for both models. These findings indicate that DyT is struggling to fully replace BN in these classic ConvNets. We hypothesize this could be related to BN layers being more frequent in these ConvNets, where they appear once with every weight layer, but LN only appears once per several weight layers in Transformers.


\end{document}